\documentclass[letterpaper, 10 pt, conference]{ieeeconf}  

\IEEEoverridecommandlockouts                              
\usepackage{cite, epsfig, mathptmx, times, amssymb, amsmath, color}
\usepackage{graphicx,booktabs,soul,color,graphics,epsfig,mathptmx,moreverb,ifpdf,cancel,framed,dblfloatfix}
\usepackage{gensymb,subcaption,graphicx,url, multirow}
\usepackage{pbox}
\usepackage{adjustbox}
\usepackage[linesnumbered,ruled,vlined]{algorithm2e}

\usepackage[x11names,dvipsnames,table]{xcolor} 
\usepackage{colortbl}
\usepackage{floatrow}
\usepackage{float}
\usepackage{caption}
\usepackage{balance}
\usepackage{xcolor}
\usepackage{amsmath,amssymb,amsfonts}
\usepackage{textcomp}

\DeclareMathAlphabet{\mathpzc}{OT1}{pzc}{m}{it}

\usepackage[firstpage]{draftwatermark}
\SetWatermarkText{This paper has been accepted for publication at IROS 2019 workshop "Challenges in Vision-based Drones Navigation". @IEEE}
\SetWatermarkVerCenter{2cm}
\SetWatermarkAngle{0}
\SetWatermarkFontSize{8pt}
\SetWatermarkColor[gray]{0.4}
\SetWatermarkScale{0.9}

\title{\LARGE \bf
Toward Underground Localization: Lidar Inertial Odometry Enabled Aerial Robot Navigation
}

\author{Jiun Fatt Chow$^1$, Basaran Bahadir Kocer$^1$, John Henawy$^2$, Gerald Seet$^1$, Zhengguo Li$^3$,  Wei Yun Yau$^3$, \\ Mahardhika Pratama$^4$
\thanks{$^{1}$The authors are with the School of Mechanical and Aerospace Engineering, Nanyang Technological University, 50 Nanyang Avenue, Singapore. {\tt\small \{jchow011,koce0001,mglseet\}@ntu.edu.sg}}
\thanks{$^{2}$John Henawy is with the School of Mechanical and Aerospace Engineering,
       Nanyang Technological University, Singapore, 639798 and Institute for Infocomm Research, A-STAR, Singapore, 138632 and is a recipient of the A*STAR SINGA scholarship.
        {\tt\small johnfari001@e.ntu.edu.sg}}%
\thanks{$^{3}$Zhengguo Li and Wei Yun Yau are with the Institute for Infocomm Research, A-STAR, Singapore, 138632
        {\tt\small \{ezgli,wyyau\}@i2r.a-star.edu.sg}}%
\thanks{$^{3}$Mahardhika Pratama is with the School of Computer Science and Engineering, Nanyang Technological University, 50 Nanyang Avenue, Singapore.
        {\tt\small mpratama@ntu.edu.sg}}%
}

\begin{document}

\maketitle
\thispagestyle{empty}
\pagestyle{empty}

\begin{abstract}
Localization can be achieved by different sensors and techniques such as a global positioning system (GPS), wifi, ultrasonic sensors, and cameras. In this paper, we focus on the laser-based localization method for unmanned aerial vehicle (UAV) applications in a GPS denied environment such as a deep tunnel system. Other than a low-cost 2D LiDAR for the planar axes, a single axis Lidar for the vertical axis as well as an inertial measurement unit (IMU) device is used to increase the reliability and accuracy of the localization performance. We present a comparative analysis of the three selected laser-based simultaneous localization and mapping (SLAM) approaches:(i) Hector SLAM; (ii) Gmapping; and (iii) Cartographer. These algorithms have been implemented and tested through real-world experiments. The results are compared with the ground truth data and the experiments are available at \url{https://youtu.be/kQc3mJjw_mw}.  
\end{abstract}

\section{Introduction}

UAVs have become a valuable platform for specific tasks such as inspection, mobile manipulation, surveillance, aerial photography and mapping due to their dexterity while flying \cite{kocer2019inspection,mahmoud2014linear}. In 2017, the Land Transport Authority (LTA) in Singapore has researched intensively on the use of UAV technology for more efficient and flexible tunnel inspections \cite{lta_drones}. To perform these tasks, it is possible to leverage model-based \cite{kocer2019aerial} and/or model-free control approaches \cite{hady2019real}. However, the localization plays a more crucial role in determining the current position and the orientation of the aerial robot with respect to a reference frame. In recent years, various algorithmic approaches have been presented for the mobile robots' localization, combining different sensor configurations, such as visual-based localization such as a stereo camera and optical flow sensor, laser-based localization and GPS \cite{Skog2016}. Outdoor localization can be easily achieved using GPS while indoor localization mainly relies on a stereo camera, ultra-wideband technology, radio waves or beacons configuration \cite{hening20173d,zhen2019estimating}. However, these methods have difficulties for navigation in a tunnel environment due to the following reasons: poor light condition, featureless environment, echo interference, and GPS-denied. 

For a potential localization in a tunnel environment, simultaneous localization and mapping (SLAM) technique can be a viable option. Some available approaches aim to explore the use of thermal and visual camera information \cite{khattak2019visual}. With an onboard illumination and semi-autonomous setting, the use of laser scanners together with cameras are used in \cite{ozaslan2017autonomous}. In a similar setting, the combination of stereo cameras and laser scanners is proposed recently in \cite{quenzel2019autonomous} for a chimney inspection. Specifically, SLAM using cameras is referred to visual-based SLAM (vSLAM) which is based on visual information only while SLAM using the LIDAR sensor is referred to as laser-based SLAM which relies on laser scan information \cite{Taketomi2017}. The laser-based SLAM technique might be superior to vSLAM in an indoor environment (e.g. deep tunnel system) where the ambient light condition is not optimum. Hence, this paper presents a comparison of potential laser-based SLAM techniques considering our desired tolerance of 20cm with a two-dimensional (2D) LIDAR sensor as the main perception input. For the vertical axis, the system is endowed by TFMini, which is a 1D Lidar sensor. An onboard IMU is used to further improve both reliability and accuracy of pose estimation \cite{henawy2019accurate} by eliminating unusable laser scan (caused by rolling and pitching). 

Three potential approaches, namely Hector SLAM, Gmapping, and Cartographer, are implemented and configured to be tested on the UAV platform. The mapping and localization performances are then compared with ground truth data in the motion capture lab, and the pose estimation error of both approaches are evaluated and discussed in the paper. The evaluations from this study might potentially determine which approach is more suitable for aerial robot localization in an indoor environment such as a deep tunnel system.

\section{Preliminaries}

\subsection{Localization and Mapping}

Localization of a mobile robot is required to determine the pose information with various sensor configurations within an environment based on an algorithm. For example, LIDARs, ultrasonic sensors, stereo cameras are common configurations for the localization. For autonomous systems, SLAM is introduced as the most widely researched topic. It can be useful for creating and updating maps within an unknown environment, while keep tracking the position of the mobile robot instantaneously. Therefore, there has been extensive research into SLAM algorithms, with reliably working solutions for typical indoor and outdoor environments using particle filters as in Gmapping. Most of them are being available as open software for individual and collaborative development \cite{Santos2013,Li2014,Kohlbrecher2011,Tardioli2014,Nikolov2017,Peel2018}.

\subsection{LIDAR Selection}

LIDAR is a remote sensing method that uses light in the form of a pulsed laser beam to measure ranges (variable distance). The capabilities of the LIDAR sensor are critical in our project since it serves as the main sensing unit for the localization algorithm. Therefore, some potential LIDAR sensors are shortlisted in Table \ref{tab:lidar}, presenting their characteristics including detecting range, the field of view (FoV), and scanning frequency.
%
%
\begin{table}[t!]
\centering
\caption {Specifications of Selected Lidars.}
	\rowcolors{2}{}{Wheat1}
	\small
	\tabcolsep=0.1cm
	\label{tab:lidar}
	\footnotesize
	\begin{tabular*}{\textwidth}{p{0.2\textwidth}p{0.2\textwidth}p{0.2\textwidth}p{0.25\textwidth}}
		\toprule
		\textbf{Model}     & \textbf{Range} (m) & \textbf{FoV} (deg) & \textbf{Frequency} (Hz) \\
		\rowcolor[rgb]{ .867,  .922,  .969}  RPLidar A1 & [0.15, 12] & 360 & 10  \\
		RPLidar A2 & [0.20, 12] & 360 & 10  \\
		\rowcolor[rgb]{ .867,  .922,  .969}  LDS-01 & [0.12, 3.50] & 360 & 5  \\
		Hokuyo URG-04LX & [0.06, 4] & 240 & 10  \\
		\rowcolor[rgb]{ .867,  .922,  .969}  Hokuyo UTM-30LX & [0.10, 30] & 270 & 36  \\
		\bottomrule	
	\end{tabular*}
\end{table}

In this study, an omnidirectional laser scan is desired to detect the features in a tunnel environment. The detecting range of the sensor must be able to reach the width of the sewerage tunnel with 6m wide, similar to the case in \cite{tan2018smart}. After some comparisons and considerations, RPLidar A1 was selected because it is a low-cost 360-degree laser scanner with a scanning rate of 10Hz.

\section{Potential 2D SLAM Techniques}

This section discusses the characteristic of chosen SLAM techniques, the configuration and fine-tuning of them to perform seamlessly with our LIDAR platform. To accomplish this goal, a personal laptop is used to perform software implementation, with the following specifications: (i) Intel Core i5-4210U CPU@1.7GHz quad-core; (ii) 8GB RAM; and (iii) NVIDIA 840M GPU.

\subsection{Hector SLAM}
Hector SLAM\footnote{\url{www.ros.org/wiki/hector\_slam}}   incorporates with 2D LIDAR sensor to generate a map from the laser scan. In contrast to other SLAM techniques (e.g. Gmapping), Hector SLAM does not require any auxiliary odometry sensor (e.g. wheel encoders) which directly measures the travel distance of a land-based robot, but only relies on the information from the laser scan matching approaches. Therefore, the Hector SLAM is more suitable for aerial vehicles. The Hector SLAM takes advantage of the low distance measurement noise and high sampling rates of LIDAR for a fast scan-matching method \cite{Kohlbrecher2011}. Another advantage of the Hector SLAM is its capability to generate multi-resolution grid maps to avoid singularity during scan matching.

A map can be generated by the Hector SLAM according to the endpoints of the laser beams hit onto the walls. Then, the transformation of the current scan is determined by the Gauss-Newton approach, which finds the best alignment of the current scan to the map generated previously.


\subsection{Gmapping}
Gmapping\footnote{\url{ www.ros.org/wiki/gmapping}} is a laser-based SLAM algorithm, which uses a Rao-Blackwellized Particle Filter SLAM approach. It is one of the most widely used SLAM methods in robotics, especially for land-based mobile robots. In general, the particle filter family of the algorithm requires high sampling particles to obtain accurate results, therefore it might have relatively increased computational complexity. Also, the depletion problem associated with this method decreases the algorithm accuracy. This arises when the elimination of a large number of particles from a sample set during the resampling step. In this context, an adaptive resampling technique has been developed to minimize the depletion problem since the resampling process is only performed when it is required. Moreover, this approach takes into account not only the movement of the mobile robot but also the most recent sensor observation with odometry information; therefore decreasing the uncertainty for the robot’s pose in the particle filter's prediction step. As a result, the number of particles required is significantly reduced since the uncertainty is lower, due to the quality of the laser scan matching process. In our experiment, the number of particles used is set to the default value of 30.

\subsection{Cartographer}
Cartographer\footnote{\url{www.ros.org/wiki/cartographer}} is an active approach that provides real-time SLAM in 2D and 3D across multiple platforms and sensor configurations. It is an open-source library, developed by Google since 2016, which is also a state of art algorithm. Worth to mention, Google Cartographer does not require a particle filter algorithm for mapping. It overcomes the issue of error accumulation during long iterations by pose estimation against a recent submap.

In 2D SLAM, the Cartographer supports running the correlative scan matcher, which is used for finding loop closure constraints with a submap (at the best-estimated position) referred to as frames. In detail, scan matching occurs at a recent submap, therefore it only depends on the recent scans. After each submap is finished, there are no longer new scans that could be inserted; it automatically checks all submaps and scans for the loop closure. A scan matcher starts to find the scan in the submap if the scans and the submaps are close enough based on the current pose estimates \cite{Yagfarov2018}.

The conversion process from a scan into a submap is given in \cite{Hess2016}. The generated submaps are presented in the form of a probability grid point which contains all the endpoints of beams that are closest to that grid point. Whenever a scan is inserted into the probability grid; the hits and misses are computed. Cartographer uses the Ceres scan matching approach to increase the accuracy of the scan pose in the submap.

\section{Results and Discussion}

All three potential SLAM techniques are implemented and tested through offline experiments. A hand-held experimental platform of the LIDAR sensor and IMU device is designed and the data collection is conducted to obtain the results for each approach. The indoor experiments are conducted in the Motion Analysis Laboratory which equipped with OptiTrack Motion Capture (Mocap) systems that allow vehicles to navigate with a less than centimeter accuracy. Mocap uses 8 off-board cameras to identify vehicle pose information (position and attitude) in a 3D space. It is an external system that tracks the position of reflective markers and provides its pose at 240 Hz. The ground truth data are received and recorded for the comparison.


There are two trajectories with two different speeds (normal walking speed and fast walking speed), with the starting point set as the origin of the coordinates system (center location), as described below.
\subsubsection{Straight line trajectories}
Starting from the origin, moving straight and then follow the rectangular path until it ends at the starting point. The heading of the LIDAR sensor is purposely remained facing forward (X-direction), thus avoid the potential noise due to the large yawing angle.
\subsubsection{8-shape trajectories}
Starting from the origin, moving along figure 8 with the heading aligned with the trajectory, ending at the starting point. 
%
\subsection{Performance Analysis of Selected SLAM Approaches} 
A group of typical localization results in the indoor experiment is illustrated from the following aspects: (i) robot path; and (ii) yaw angle. 
\subsubsection{Normal moving speed}
\begin{figure}[!b]
    \centering
    \includegraphics[width=\textwidth]{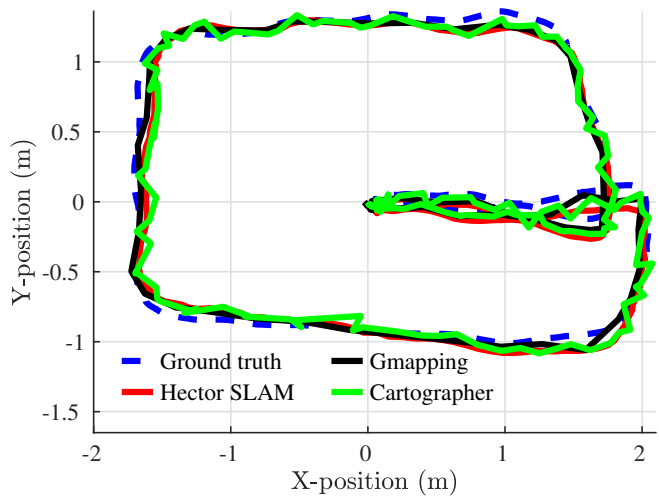}
    \caption{UAV path comparison: straight lines in nominal velocity trajectories.}
    \label{fig:l1}
\end{figure}

\begin{figure}[!b]
    \centering
    \includegraphics[width=\textwidth]{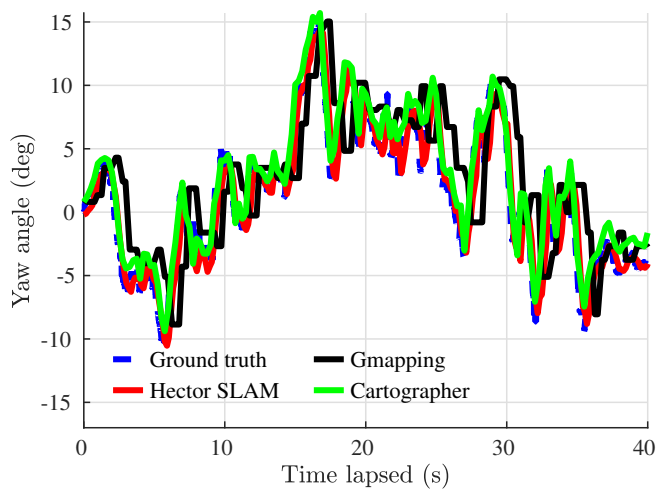}
    \caption{Yaw angle comparison in straight lines for nominal velocity trajectories.}
    \label{fig:l2}
\end{figure}

Firstly, the results from the first case are illustrated in Fig. \ref{fig:l1} and Fig. \ref{fig:l2}. All trajectories are plotted on the same graph to have a clearer comparison of the selected algorithms. In Fig. \ref{fig:l1}, the localization results from the Hector SLAM and Gmapping are comparably smoother than the outcome of the Cartographer which shows fluctuation and jerkiness. This result is possibly due to the characteristics of the Cartographer that fuses multiple sensor data, but there are only LiDAR sensors in our case. On the other hand, it is difficult to conclude that either the Hector SLAM or the Gmapping is more accurate only by this test.

According to our observation, the localization performance of the laser-based algorithm is affected by three variables, namely: the performance of the LIDAR sensor, the feature extraction within the environment and the matching algorithm.

From Fig. \ref{fig:l2}, a significant time delay is observed in the Gmapping algorithm, which also exhibits slightly discrete movement. This behavior is mainly because of the slower updating frequency of the Gmapping algorithm as compared to others. On the other hand, both Hector SLAM and Cartographer give reliable positioning results, but the latter algorithm shows fluctuation during each peak.

\begin{figure}[!t]
    \centering
    \includegraphics[width=\textwidth]{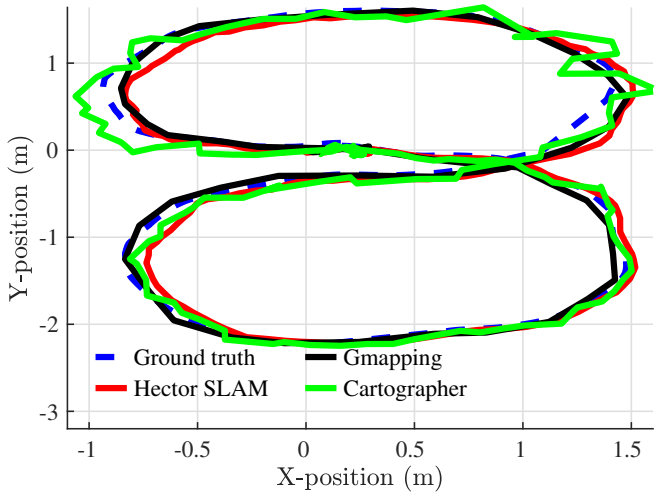}
    \caption{UAV path comparison: an 8-shape in nominal velocity trajectories.}
    \label{fig:81}
\end{figure}

\begin{figure}[!t]
    \centering
    \includegraphics[width=\textwidth]{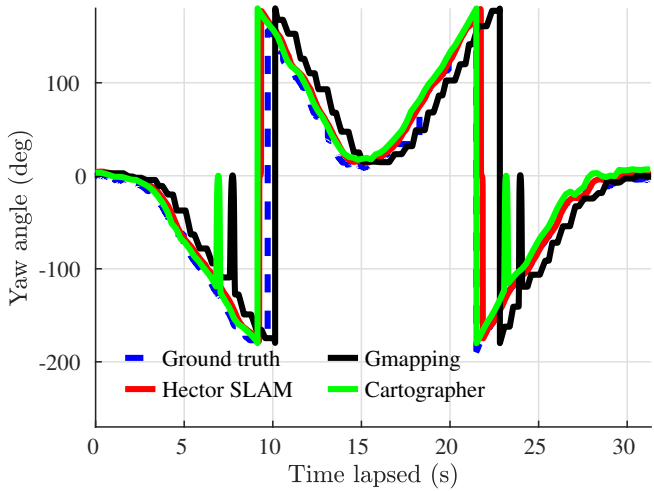}
    \caption{Yaw angle comparison in an 8-shape for nominal velocity trajectories.}
    \label{fig:82}
\end{figure}

In the second case, the circular path is used to test the robustness of each SLAM technique when facing large rotating in the yaw angle. The results are shown in Fig. \ref{fig:81} and Fig. \ref{fig:82}. Once again, Hector SLAM had better results as compared to the Cartographer which is wavy and the Gmapping which has a high time delay. Notably, the time shift of the Gmapping could be larger when the simulation time increases. We can also observe that there are two sudden peaks of the yaw estimation from both the Gmapping and the Cartographer packages.
\subsubsection{Fast moving speed}
\begin{figure}[t!]
    \centering
    \includegraphics[width=\textwidth]{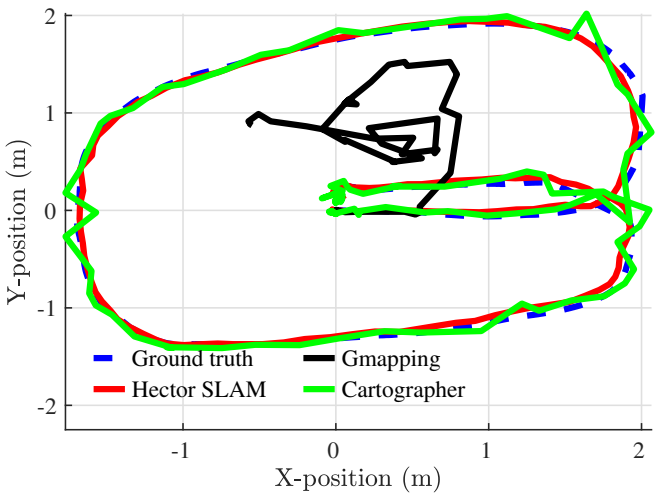}
    \caption{UAV path comparison: straight lines in faster velocity trajectories.}
    \label{fig:ll1}
\end{figure}

\begin{figure}[t!]
    \centering
    \includegraphics[width=\textwidth]{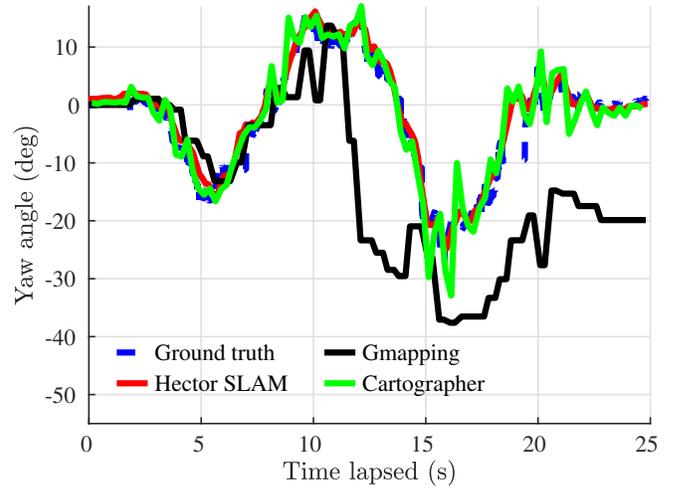}
    \caption{Yaw angle comparison in straight lines for faster velocity trajectories.}
    \label{fig:ll2}
\end{figure}

\begin{figure}[t!]
    \centering
    \includegraphics[width=\textwidth]{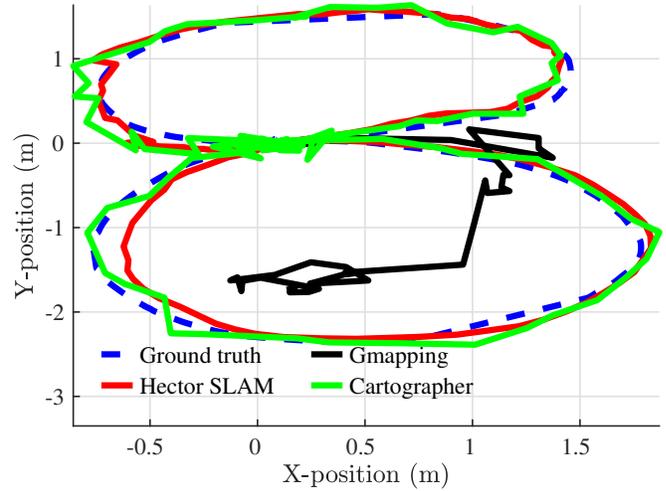}
    \caption{UAV path comparison: an 8-shape in faster velocity trajectories.}
    \label{fig:881}
\end{figure}

A similar experiment is conducted with a faster speed to test the robustness of each algorithm. As can be seen in Fig. \ref{fig:ll1}, the Gmapping is not functioning properly. Hector SLAM has the best pose estimation where else the Cartographer generated a wobbly path. In Fig. \ref{fig:ll2}, the ground-truth data update stops at some instants in the Mocap system due to the communication problems.
During the faster circular trajectory shown in Fig. \ref{fig:881}, the Gmapping has failed to perform SLAM properly while the Cartographer generated a fluctuated data. Only the Hector SLAM’s trajectory has the closest trajectory as compared to the ground truth values.
\begin{figure}[t!]
    \centering
    \includegraphics[width=\textwidth]{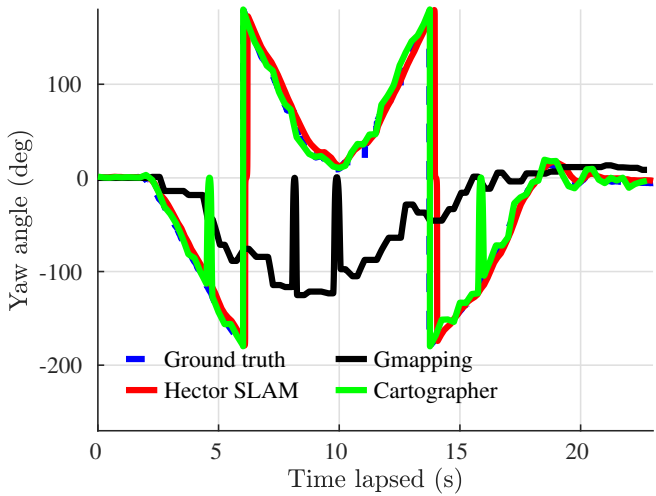}
    \caption{Yaw angle comparison in an 8-shape for faster velocity trajectories.}
    \label{fig:882}
\end{figure}

Notably, the Cartographer experienced a sudden spike twice in determining the heading of our robot when the system rotates -120 degrees of yaw, shown in Fig. \ref{fig:882} (green line). There are also some stationary instants from ground-truth value during the experiment, this might be due to the blockade of reflective markers during the operation.

\subsection{Further Analysis of SLAM Performance}
To further analyze the results, we use the root mean square formula (Eq. 1) to determine the accuracy of each approach compared to the ground truth.
$$
 RMSE = \sqrt{\frac{1}{n}\Sigma_{i=1}^{n}{\Big(d_t - d_e\Big)^2}} \eqno{(1)}
$$
where $d_{t}$ is the true displacement and $d_{e}$ is the estimated displacement. 

To calculate the displacement between the true pose and the estimated pose:
$$
 d = \sqrt{\Big(x_t - x_e\Big)^2 + \Big(y_t - y_e\Big)^2} \eqno{(2)}
$$
where the subscript $t$ denotes the truth data, and the subscript $e$ denotes the estimated data. Henceforth, some conclusions on each technique are shown in Table \ref{tab:overall}.
%
%
%
\begin{table}[b!]
	\rowcolors{2}{}{Wheat1}
	\centering
	\small
	\tabcolsep=0.1cm
	\caption {Comparison of Different Approaches: RMSE Values in cm.} \label{tab:overall}
	\footnotesize
	\begin{tabular*}{\textwidth}{p{0.2\textwidth}p{0.15\textwidth}p{0.15\textwidth}p{0.15\textwidth}p{0.15\textwidth}}
		\toprule
		\textbf{Approach}     & \textbf{Linear Nominal} & \textbf{Circular Nominal} & \textbf{Linear Fast} & \textbf{Circular Fast} \\
		\rowcolor[rgb]{ .867,  .922,  .969}  Hector SLAM & 9.39 & 14.83 & 11.47 & 24.69  \\
		Gmapping & 40.10 & 42.60 & 133.95 & 196.31  \\
		\rowcolor[rgb]{ .867,  .922,  .969}  Cartographer & 16.70 & 14.94 & 15.95 & 24.56  \\
		\bottomrule	
	\end{tabular*}
\end{table}
Firstly, the Hector SLAM algorithm relies only upon the laser scan matching without the use of odometry, which could be an advantage for our aerial robot platform. Apart from that, the Hector SLAM also provides accurate pose estimation, with an average error of 15.09cm in translation after taking average RMSE of all scenarios. On the other hand, the Gmapping shows its robustness only in slow motion situations, but the time delay accumulated over time. Most importantly, the Gmapping is malfunctioning in faster movements. Lastly, the Cartographer achieves the fastest computation time and had a decent average accuracy of 18.04cm in translation. Worth to mention, the Cartographer is designed for multiple sensors platform, therefore it is expected to obtain more accurate results in a multi-sensor based system.

\subsection{On-board Experiments}
\begin{figure}[t!]
    \centering
    \includegraphics[width=\linewidth]{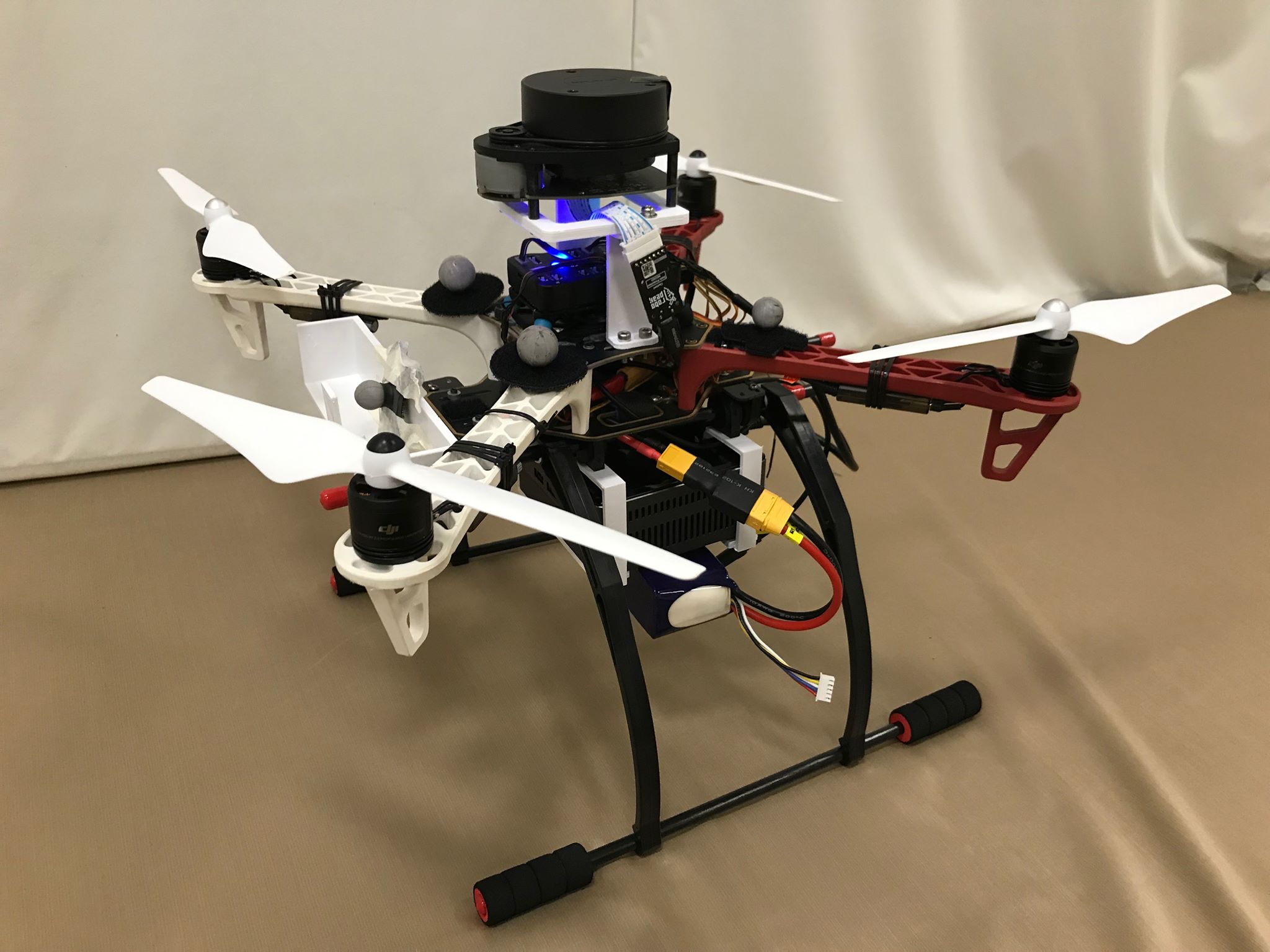}
    \caption{Aerial robot platform.}
    \label{fig:onboard}
\end{figure}

From the previous off-board experiments, the Hector SLAM is selected as an optimum SLAM algorithm considering our limitations for the sensor instrumentation. Except for 2D Lidar and onboard IMU, an additional distance sensor is needed to provide information for the altitude. Since it is tiny, low cost and consumes low power with a detecting range of 0.30m - 12m, TFmini Lidar is selected. It is configured with QGroundControl. Our UAV system is shown in Fig. \ref{fig:onboard}, where Intel NUC serves as an on-board processing unit which receives laser scan data and utilizes the Hector SLAM algorithm meanwhile fusing the altitude (from TFmini) to generate real-time 6 DoF position information.

\begin{figure}[t!]
    \centering
    \includegraphics[width=\textwidth]{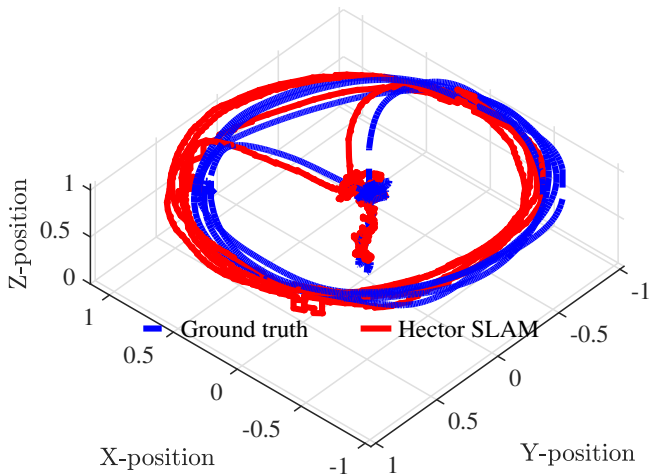}
    \caption{3D pose estimation of Hector SLAM with the ground truth comparison.}
    \label{fig:3D}
\end{figure}

During real-time experiments, 4 scenarios are carried out:
\begin{itemize}
\item Normal/fast speed circle path;
\item Normal/fast speed ‘8’ path.
\end{itemize}
The aerial robot is controlled via ROS for the desired trajectories. All the necessary nodes are running onboard. The recorded data are extracted and the displacement is calculated in 3D space, followed by the RMSE formula.
$$
 D = \sqrt{\Big(x_t - x_e\Big)^2 + \Big(y_t - y_e\Big)^2 + \Big(z_t - z_e\Big)^2} \eqno{(3)}
$$
A summary of RMS errors in 3D translations for all cases is shown in Table \ref{tab:3D}. Since the generated paths are similar to each other, only the 3D visualization of the normal speed circle path is given in Fig. \ref{fig:3D}.
\begin{table}[b!]
	\rowcolors{2}{}{Wheat1}
	\centering
	\small
	\tabcolsep=0.1cm
	\caption {Hector SLAM: Comparison of Different Trajectories.} \label{tab:3D}
	\footnotesize
	\begin{tabular*}{\textwidth}{p{0.48\textwidth}p{0.48\textwidth}}
		\toprule
		\textbf{Trajectories}     & \textbf{RMSE} (cm)  \\
		\rowcolor[rgb]{ .867,  .922,  .969}  Circular Nominal & 19.08   \\
		Circular Fast & 18.85   \\
		\rowcolor[rgb]{ .867,  .922,  .969}  8-shape Nominal & 17.27  \\
		8-shape Fast & 17.69   \\
		\bottomrule	
	\end{tabular*}
\end{table}
%
%

Notably, the detecting range of TFmini is only from 0.30 to 12m, therefore any distance below 30cm will be considered as a minimum value of 0.30m. Since the placement of TFmini was 8cm offset below the CG of the drone, the effective detecting range of altitude is 0.38m to 12.08m. Henceforth, the limitation of this fusion method is the blind zone when altitude is under 38cm.

In summary, the 3D pose information is obtained with the fusion of the Hector SLAM and the TFmini sensor. According to the ground truth data, the RMSE has stayed below 20cm for all the cases. 

\section{Conclusion}
In this work, three selected approaches were implemented on the aerial robot system endowed by a 2D Lidar for horizontal axes and 1D Lidar for the vertical axis. A set of offline experiments were conducted in different velocities and conditions to measure their tracking and mapping performances. From the results, it was concluded that the Hector SLAM package obtained a reasonable localization performance. Also, the on-board experiments showed that it was achieved to keep the RMSE below 20cm in 3D translation. At the same time, the Cartographer package is also preferable due to its potential with fusing different perception units. 

In our future work, we intend to carry out the experiments in a real tunnel environment to obtain the actual environmental conditions. Some difficulties can be expected, for example, the rough surface and the humidity within a deep tunnel might not be optimum for a laser sensor. As a solution, the robustness of the laser-based algorithm can be improved by fusing multiple SLAM algorithms such as a visual-based SLAM that using a stereo camera to detect the features within an environment. This would potentially achieve the goal of autonomous UAV inspection in a deep tunnel system without human intervention. Moreover, different lidars (e.g., Velodyne and Hokuyo) are considered to be used in our research for further improvement.  

\addtolength{\textheight}{-1cm}   

\balance
\bibliography{bib,IEEEabrv}
\bibliographystyle{IEEEtran}

\end{document}